\documentclass{article}

\usepackage{microtype}
\usepackage{graphicx}
\usepackage{subfigure}
\usepackage{booktabs} 

\usepackage{amsthm}
\usepackage{multirow}
\usepackage{graphicx}
\usepackage{amssymb}
\usepackage{amsfonts}
\usepackage{amssymb}
\usepackage{amsmath}
\usepackage{wrapfig,lipsum,booktabs}
\usepackage{float}
\usepackage{pdfpages}
\usepackage{graphicx}
\graphicspath{ {./figures/} }
\usepackage{algorithm}
\usepackage{algpseudocode}
\algrenewcommand\algorithmicrequire{\textbf{Input:}}
\algrenewcommand\algorithmicensure{\textbf{Output:}}
\usepackage{multirow}

\newtheorem{theorem}{Theorem}
\newtheorem{claim}{Claim}

\usepackage{commath}
\usepackage[english]{babel}

\usepackage[utf8]{inputenc} 
\usepackage[T1]{fontenc}    
\usepackage{hyperref}       
\usepackage{url}            
\usepackage{booktabs}       
\usepackage{amsfonts}       
\usepackage{nicefrac}       
\usepackage{microtype}      
\usepackage{xcolor}         
\usepackage{enumitem}

\newcommand{\name}{FLuRKA}

\usepackage[preprint]{neurips_2023}

\usepackage[utf8]{inputenc} 
\usepackage[T1]{fontenc}    
\usepackage{hyperref}       
\usepackage{url}            
\usepackage{booktabs}       
\usepackage{amsfonts}       
\usepackage{nicefrac}       
\usepackage{microtype}      
\usepackage{xcolor}         



\title{FLuRKA: Fast and accurate unified Low-Rank \& Kernel Attention}

\author{%
Ahan Gupta \\
University of Illinois Urbana-Champaign \\
\texttt{ag82@illinois.edu}
\And
Hao Guo \\
University of Illinois Urbana-Champaign \\
\texttt{haoguo2@illinois.edu}
\And
Yueming Yuan \\
University of Illinois Urbana-Champaign \\
\texttt{yy28@illinois.edu}
\And
Yanqi Zhou \\
Google DeepMind \\
\texttt{yanqiz@google.com}
\And
Charith Mendis \\
University of Illinois Urbana-Champaign \\
\texttt{charithm@illinois.edu}
}

\begin{document}

\maketitle

\begin{abstract}
Many efficient \textit{approximate} self-attention techniques have become prevalent since the inception of the transformer architecture. Two popular classes of these techniques are low-rank and kernel methods. Each of these methods has its strengths. We observe these strengths synergistically complement each other and exploit them to fuse low-rank and kernel methods, producing a new class of transformers: \name{} (\textbf{F}ast \textbf{L}ow-\textbf{R}ank \& \textbf{K}ernel \textbf{A}ttention). \name{} are highly \textit{training-efficient} with faster model speeds \textit{and} similar model qualities compared to constituent low-rank and kernel methods. We theoretically and empirically evaluate the speed and quality of \name{}. Our model speed analysis posits a variety of parameter configurations where \name{} exhibit speedups over low-rank and kernel approximations and our model quality analysis bounds the error of \name{} with respect to full-attention. Empirically, we instantiate three \name{} variants which experience speedups of up to 3.3x and 1.7x over low-rank and kernel methods respectively. This translates to speedups of up to 20x over models with flash-attention. Across a diverse set of tasks spanning language modeling, language understanding, long sequence modeling, machine translation, and image classification, \name{} achieve comparable accuracy with underlying low-rank and kernel approximations, occasionally surpassing both.  
\end{abstract}

\section{Introduction}
Transformers have been widely adopted across various domains, powering popular applications like ChatGPT, Gemini Pro, and Claude, which handle millions of queries per day \cite{scale}. To effectively train \textit{and} deploy high-quality models at this scale, transformers must be \textit{training efficient}. This entails balancing two crucial factors. (1) \textit{Model speed}, which corresponds to models with low step-times and fewer FLOPs per step. (2) \textit{Model quality}, which corresponds to expressive models that yield low losses with few training tokens. Highly training-efficient transformers are fast and of high quality. However, achieving both simultaneously is challenging, as higher quality transformers often require larger parameter counts and data-set sizes \cite{sample-efficient, chinchilla}, leading to slower model speeds. This trade-off is further exacerbated by the quadratic dependence in run-time on \textit{increasing} input sequence lengths (32k for GPT-4, 100k for Claude, and 1.5M for Gemini-pro \cite{gpt4, claude, gemini}). 

To design training-efficient transformers, researchers have developed linear time self-attention approximations, falling into three main categories: sparse \cite{sparse-transformer, reformer, big-bird}, low-rank \cite{linformer}, and kernel \cite{performer, rnns, random-features-attention, eva, lara} methods. Sparse methods compute a subset of the full-attention matrix, low-rank methods leverage the low-rank property of self-attention and kernel methods approximate the softmax kernel. While each category excels in specific tasks – sparse methods in document retrieval, low-rank methods in long-sequence modeling, and kernel methods in classification – their high model quality is limited to these narrow domains \cite{lra}.

\begin{wraptable}[16]{R}{0.40\linewidth}
\centering
\scalebox{0.85} {
\begin{tabular}{c|l|c|c}
  \toprule
  & Model & Loss & EFLOPs \\
  \midrule
  & Full Attention & 0.036 & 35.9   \\
  \midrule
  \multirow{3}{*}[0.4ex]{\rotatebox[origin=c]{90}{\textbf{\name{}}}} & Linformer-Perf. & \multirow{3}{*}{0.036} & 24.5 \\
  & Linformer-RNN &  & 26.2 \\
  & Linformer-EVA & & 24.2\\
  \bottomrule
\end{tabular}}
 \caption{We compare the validation loss (model quality) against FLOPs count (model speed) of \name{} variants and full-attention on the auto-regressive English to German machine-translation benchmark. EFLOPs are ExaFLOPs ($10^{18}$).}
 \label{table:isoflops-motivation}
\end{wraptable}

To design training-efficient models across \textit{diverse} tasks, researchers have experimented with unifying different linear time self-attention approximation techniques to combine their individual strengths. Scatterbrain (SB) \cite{scatterbrain} unifies kernel (K) and sparse (S) methods by computing an attention matrix whose values come from S if the cosine similarity between a query and key is large, else comes from K. Though expressive, SB requires K and S to be entirely computed, resulting in $T(SB) > \min(T(K), T(S))$ due to parallelism, where $T$ is the step-time (model speed) of a corresponding method. Long-short (LS) \cite{longshort} unifies low-rank (LR) and sparse (S) methods by computing low-rank approximation and concatenating the attention matrix with a sparse local window of keys and values for each query. Though expressive, LS also requires the entire computation of LR and S to be computed, similarly resulting in $T(LS) > \min(T(LR), T(S))$. Existing fusions are limited in their training-efficiencies due to model speeds lower-bounded by the faster of their constituent methods. Constructing a unification faster than both constituent models requires an additional approximation that \textit{partially computes} at least one of them. However, naively doing so may reduce model expressivity and adversely impact quality. 

In this work, we bridge this gap by proposing a unification of low-rank (LR) and kernel (K) methods, \name{} (\textbf{F}ast and accurate \textbf{L}ow-\textbf{r}ank and \textbf{K}ernel \textbf{A}ttention), a new class of high-quality unified transformers that is faster than its constituent models. We specifically construct a unification that exploits the orthogonal compute strengths of low-rank methods (efficient linear transformations) and kernel methods (efficient softmax kernel computations), resulting in model speeds faster than $\min(T(LR), T(K))$. Additionally, we demonstrate our construction tightly approximates full attention and is of high-quality. Empirically, we validate \name{} and show they are up to 3.3x and 1.7x faster than constituent low-rank and kernel approximations with comparable model qualities across 6 diverse data-sets spanning image and text modalities. \name{} variants are highly training-efficient, requiring fewer FLOPs than full attention to achieve similar loss levels (see table \ref{table:isoflops-motivation}). This significantly reduces the computational cost of training high-quality models for a diverse range of tasks. Specifically, we make the following contributions:
\begin{itemize}
    \setlength\itemsep{6pt}
    \item A technique to unify two classes of approximations: low-rank and kernel methods, to produce a new class of transformers, \name{}. \name{} are faster than constituent low-rank and kernel methods, incurring step-times (model speeds) faster than $\min(T(LR), T(K))$ \textit{and} are of high quality, resulting in high training-efficiencies. 
    \item A theoretical analysis of \name{} where we delineate precisely when \name{} are faster than low-rank and kernel methods and additionally show that \name{} have tight approximation error with respect to full-attention.
    \item Empirical studies on three different instantiations of low-rank and kernel methods demonstrating the generality of our technique. Our studies show that \name{} are up to 1.7x and 3.3x faster compared to kernel and low-rank methods, respectively. Moreover on language modeling (on wikitext-103 \cite{wikitext}), language understanding (on GLUE \cite{glue}), long sequence modeling (on LRA \cite{lra}), machine translation (on english to german and english to french) and image classification (on imagenet \cite{imagenet}), \name{} are competitive with, and occasionally surpass, the low-rank and kernel methods that compose them.
\end{itemize}
\vspace{-7pt} 

\section{Background and Related Work}
\label{background}
The backbone of the transformer is multi-head-self-attention (MHSA) \citep{full-attention}. MHSA computes the following matrix: $Concat(Head_1, Head_2, ..., Head_h)$ where $Head_i$ is: 
$$\underbrace{softmax(QW_i^Q(KW_i^K)^T)}_{A_i}VW_i^V$$ 
The matrices $Q$, $K$ \& $V$ $\in \mathbb{R}^{N \times d_m}$ are the input matrices consisting of $N$ points embedded in $\mathbb{R}^{d_m}$, where $d_m$ and $N$ are known as the embedding dimension and sequence length respectively. $W_i^Q$, $W_i^K$ and $W_i^V \in \mathbb{R}^{d_m \times d_h}$ are linear transformations. The matrix $A_i$ is known as the attention matrix and the softmax is taken row-wise in the product $QW_i^Q(KW_i^K)^T$. Self-attention is expensive due to the matrix $A_i$ being of size $O(N^2)$. 
\subsection{Efficient Approximations}
\paragraph{Low-rank Methods}
Low-rank methods exploit the observation that the matrix $A_i$ is of low-rank. Motivated by this, linformer \citep{linformer}, a SOTA low-rank technique, constructs a low-rank approximation of $Head_i$ via:  
$$Softmax(QW_i^Q(E_1KW_i^K)^T)E_2VW_i^V$$ Where $E_1$ and $E_2$ are matrices in $\mathbb{R}^{d_k \times N}$ whose entries are sampled from $N(0, \frac{1}{d_k})$. The resultant size of $A_i$ is $O(Nd_k)$, where $d_k$ is the downsampling-factor: a hyperparameter that is set prior to training. In practice, $d_k$ is agnostic of $N$ and is much smaller, linearizing self-attention.
\paragraph{Kernel Methods}
Kernel methods \citep{performer, lara, eva, rnns} replace the softmax with a cheaper approximation. They map the input to a space where the dot-product approximates the softmax, computing the following for $Head_i$: $$\phi(QW_i^Q)(\phi(KW_i^K)^TVW_i^V)$$ Where $\phi : \mathbb{R}^{d_h} \rightarrow \mathbb{R}^{d_p}$ is a kernel with the property: $softmax(x,y) \approx \phi(x)^T \cdot \phi(y)$. Kernel methods do not materialize the matrix $A_i$ directly, instead by multiplying out the matrices in an optimal manner, they linearize self-attention to $O(Nd_p^2)$. There are two categories of kernel methods: random feature and non-random-feature-mapped kernels. Random feature mapped kernels are parameterised by $m$ i.i.d random-variables $\psi_1, \psi_2, ... \psi_m$ with $\psi_i : \mathbb{R}^{d_h} \rightarrow \mathbb{R}$, where $\mathbb{E}[\psi_i(x)^T \cdot \psi_i(y)] = \exp(x^T \cdot y)$, and are defined as: $\phi(x) = \frac{1}{\sqrt{m}}[\psi_1(x), \psi_2(x), ... \psi_m(x)]$. Non-random feature mapped kernels have no element of randomness and instead use kernels like the exponential linear unit \citep{rnns}. We fuse low-rank methods with both random (performer, EVA) and non-random (RNN) feature mapped kernel methods. 

\subsection{Unified Attention Mechanisms}

Existing unified attention methods exploit the synergies between disparate attention mechanisms for higher model quality. Scatterbrain \citep{scatterbrain} unifies sparse with kernel methods whereas longshot \cite{longshort} unifies sparse with low-rank methods. However, both these existing fusions have model speeds lower-bounded by the faster of their respective constituents, limiting their training-efficiencies. Although we also propose a unification of different transformers, to the best of our knowledge we are the first to explore a unification: (1) of low-rank and kernel methods, (2) that produces high quality transformers with faster model speeds than \textit{both} constituent models.

\section{\name{}: Fused Low-Rank and Kernel Attention} 
\label{section:cmp}
Constructing unified high-quality transformers whose model speeds are faster than their constituents is challenging due to two reasons. (1) An additional approximation is required to partially compute at least one constituent method to enhance model speed. (2) This approximation cannot adversely impact the unified model's quality.

Nevertheless, we propose a unification of low-rank (LR) and kernel (K) methods that achieves both attributes. In doing so, we produce high-quality transformers whose model speeds are \textit{faster} than $\min(T(LR), T(K))$, resulting in high training-efficiencies. Moreover, our technique is general, enabling a unification of any low-rank and kernel method, producing a family of transformer models. 

\begin{wrapfigure}{r}{0.5\textwidth}
  \begin{center}
    \includegraphics[width=0.48\textwidth]{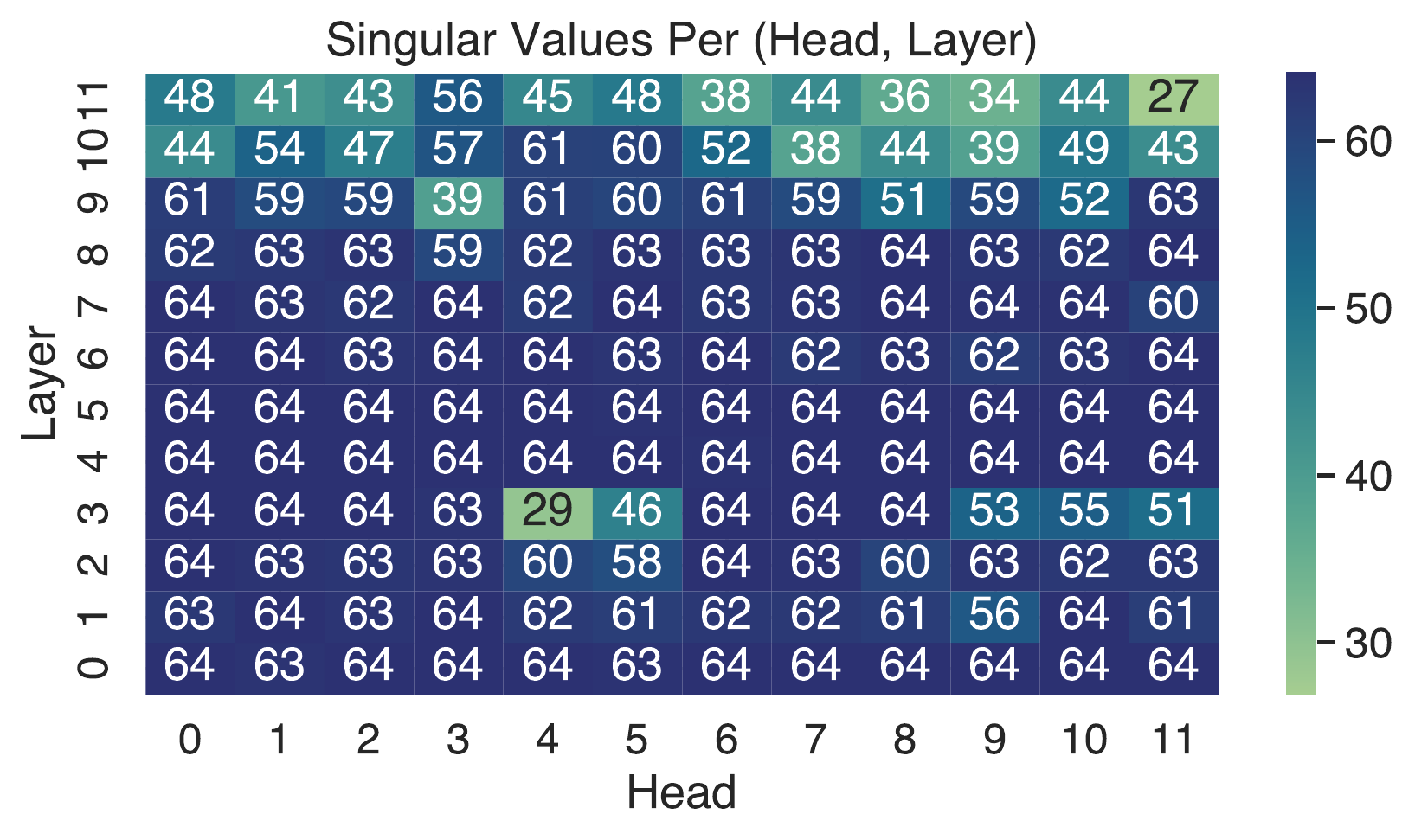}
  \end{center}
  \caption{We take 12-layer 12-head performer pre-trained on wikitext-103 and plot the number of unique singular values in the SVD of the kernelized attention matrix: $\phi(QW^Q)\phi(KW^K)$ for every alternate attention head in each layer.}
  \label{fig:svd}
\end{wrapfigure}

\textbf{Naive Unification.} We observe that the attention matrix implicitly materialized through cheaper softmax kernel approximations: $\phi(QW_i^Q)\phi(KW_i^K)$ is of low-rank. This is demonstrated in figure \ref{fig:svd}, where we compute the number of singular values for every attention-head in every alternate layer of a pre-trained performer model. The size of the attention matrix is 128x128, yet the attention-head of highest rank is 64 (the size of the head hidden dimension) with the rank of the attention-heads decreasing in later layers. 

Therefore, we can apply low-rank (LR) approximation over the kernelized (K) attention matrix to unify the two techniques. The full unification of $Head_i$ will look as follows: \begin{equation} \label{naive-fusion}
\phi(QW_i^Q)(E_1\phi(KW_i^K)(E_2VW_i^V)) 
\end{equation}
Although this maintains model expressivity, the resulting model is slow as the entire low-rank and kernel approximations need to be computed. Moreover, they cannot be parallelised, resulting in model speeds lower-bounded by $T(LR) + T(K)$. 

\begin{figure*}[t]
    \centering
    \includegraphics[width=\textwidth]{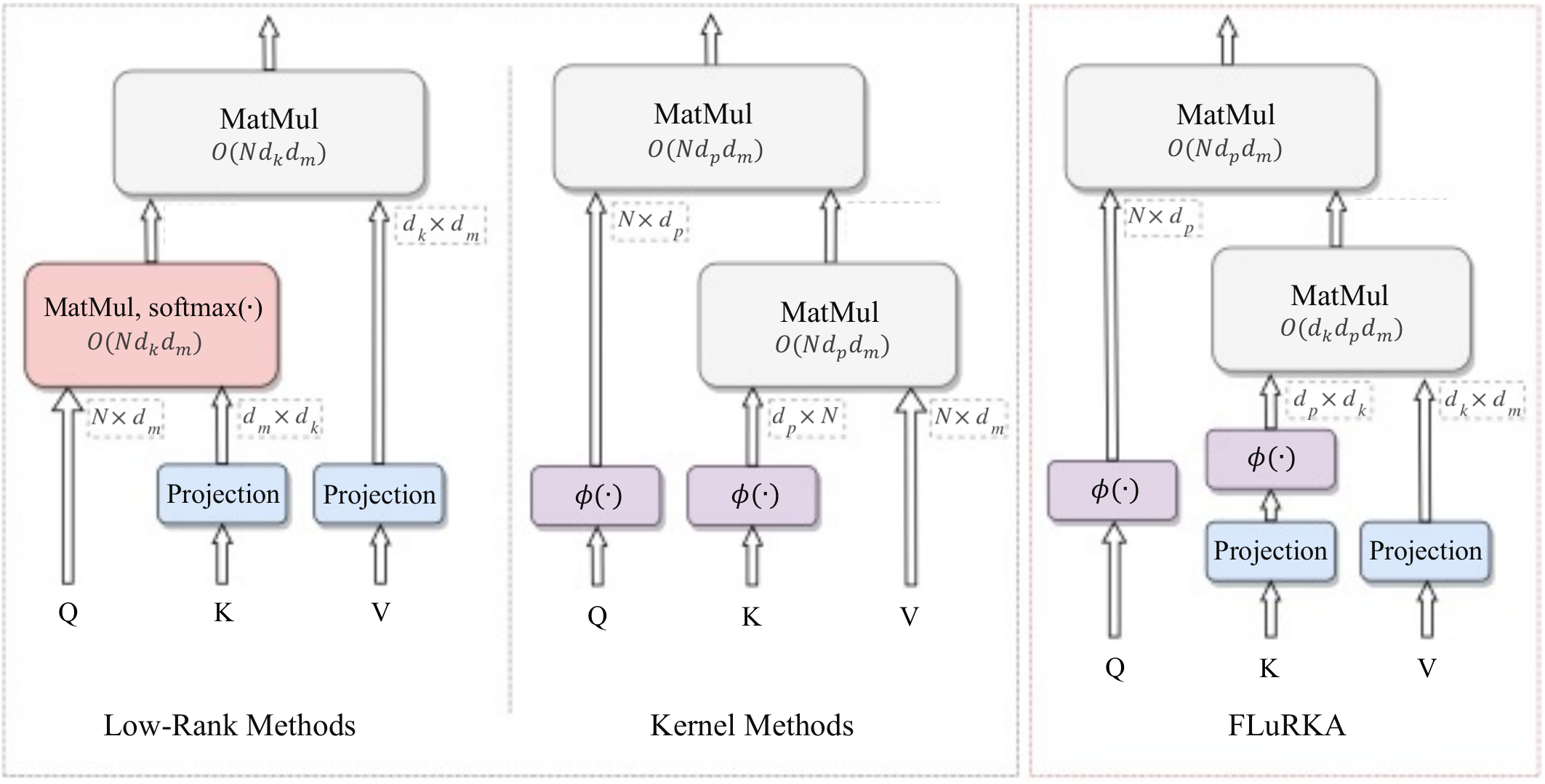}
    \caption{The pipeline of operations involved in constructing \name{}, parameterized by a kernel $\phi$, compared to low-rank and kernel methods. $N$, $d_m$, $d_k$ are the sequence length, hidden dimension, and downsampling factors, respectively. $d_p$ is the dimension of the kernelized queries and keys.}
    \label{fig:swiftformer}
\end{figure*}

\textbf{Optimized Unification.} To propose an optimized unification whose model speed is faster than equation \ref{naive-fusion} \textit{and} is of comparable model quality, we observe that low-rank and kernel methods enhance the speed of self-attention in orthogonal ways. On the one hand, low-rank methods contract the keys and values through multiplication with the $E_1$ \& $E_2$ matrices, reducing the costs of linear-transformations and $A_i(E_iVW_i^V)$ products. On the other hand, kernel methods cheaply approximate the softmax kernel through an inexpensive kernel mapping of the queries and keys, reducing the costs of softmax kernels. 

A faster unified model will therefore incur low linear-transformation and softmax approximation cost. This only occurs when the keys and values are contracted \textit{before} the linear-transformations \textit{and} softmax approximation, which requires low-rank (LR) approximation to be applied \textit{before} kernel (K) approximation, in contrast to the naive unification in equation \ref{naive-fusion}. This gives us the optimised construction of \name{} as follows: \begin{equation} \label{optimised-fusion}
    \phi(QW^Q)(\phi((E_1K)W^K)((E_2V)W^V))
\end{equation} Such a unification incurs model speeds \textit{faster} than $\min(T(LR), T(K))$ while retaining the quality of constituent models, resulting in highly training-efficient models. We theoretically analyze the speed and quality of equation \ref{optimised-fusion}. Claim \ref{claim:one} gives a tight bound delineating for what input sequence lengths \name{} incur speeds faster than $\min(T(LR), T(K))$, and theorem \ref{accuracy-claim} tightly bounds the approximation error with respect to full-attention. 

\textbf{Model speed theoretical analysis.} We theoretically analyze the model speed of \name{} and delineate precisely when \name{} are faster than their underlying low-rank and kernel approximations. 

\begin{claim}
    For sequence length: $N$, hidden dimension: $d_m$, downsampling factor: $d_k$, head hidden dimension $d_h$, number of heads $H$, when: $$N > d_k(H+2) > d_m > d_k > d_h$$ \name{} incur fewer FLOPs against both kernel and low-rank methods.
    \label{claim:one}
\end{claim}

We note that claim \ref{claim:one} is tight, resulting in faster models for input sequence lengths as short as several thousand. We further describe two additional scenarios where \name{} are faster than their constituents in Appendix B.

\textbf{Model quality theoretical analysis.} The following theorem bounds the approximation error with respect to full-attention for a \textit{generic} \name{} under reasonable assumptions.
\begin{theorem}
Suppose we have a random feature map $\phi$ defined as follows: $$\phi(x) = \frac{1}{\sqrt{m}}[\psi_1(x), \psi_2(x), ... \psi_m(x)]$$ such that: $$\mathbb{E}[\psi_i(x)^T\cdot\psi_i(y)] = \exp(x^T \cdot y)$$ Then for any $Q_i$, $K_i$, $V_i$ $\in \mathbb{R}^{n\times d_m}$ and $W_i^Q$, $W_i^K$, $W_i^V$ $\in \mathbb{R}^{d_m \times d_h}$, and $k = 5\log(d)/(\epsilon_2^2 - \epsilon_3^2)$. We have, for the matrices $E_i = \delta R, F_i = e^{-\delta}R$ where $R \in \mathbb{R}^{n \times k}$ whose entries are iid sampled from $N(0, 1/k)$ and a random feature based kernel method parameterised by $\phi$, with $\epsilon_4 > 0$: 
$$\lVert \phi(QW_i^Q)\phi(E_1KW_i^K)^TF_iVW_i^V - A_iVW_i^V \lVert_{\infty}< \epsilon_4\lVert F_iVW_i^V\lVert_{\infty} + \epsilon_1 \lVert A_i\lVert_{2} \lVert VW_i^V\lVert_{2}$$

Occurs with probability at least $1 - o(1)$ for large enough $m$.

\label{accuracy-claim}
\end{theorem}

\textbf{Proof Sketch.} Given $\mathbb{E}[\psi_i(x)^T \cdot \psi_i(y)] = \exp(x^T \cdot y)$. We can show, by the law of large numbers, that: 
\begin{equation}
\lim_{m \rightarrow \infty}\phi(x)^T \cdot \phi(y) = \exp(x^T \cdot y)
\end{equation} Which allows us to bound the error of a kernel method by stating that for an $\epsilon_4 > 0$ $\exists m \in \mathbb{Z}^{+}$: 
\begin{equation}
\lVert \hat{A_i} - A_i \lVert_{\infty} < \epsilon_4
\label{equation:err-kernel}
\end{equation} where $\hat{A_i}$ is the attention matrix materialised by a kernel method. We can then substitute $K=E_1K$ and right multiply \ref{equation:err-kernel} by $\lVert F_iVW_i^V \lVert_{\infty}$ to bound the error of a \name{} with respect to low-rank: 
\begin{equation}
\lVert \phi(QW_i^Q)\phi(E_1KW_i^K)^TF_iVW_i^V - softmax(QW_i^Q(E_1KW_i^K)^T) F_iVW_i^V\lVert_{\infty} < \epsilon_4\lVert F_iVW_i^V\lVert_{\infty}
\label{equation:swiftformer-low-rank}
\end{equation}
We then apply the triangle inequality to the sum of \ref{equation:swiftformer-low-rank} and linformer theorem 2 \citep{linformer} to recover our bound. For a detailed proof of theorem \ref{accuracy-claim}, refer to the Appendix.


\section{Evaluation}
\label{evaluation}
The design of our experiments is motivated by the following question: are \name{} empirically \textit{more} training-efficient compared to underlying low-rank and kernel methods? To answer this question, we instantiate three \name{} variants which unify linformer (low-rank approximation) with performer \citep{performer}, RNN \citep{rnns}, and EVA \citep{eva} (kernel approximation). We assess the model speed and quality of the three \name{} variants compared to underlying low-rank and kernel approximations. Section \ref{performance-evaluation} details our experiments on model speed, and section \ref{accuracy-evaluation} details our experiments on model quality. Section \ref{sub-section:ablations} details our ablation studies.  

\textbf{Experimental Setup.} All our experiments were done on a cluster of 4 A100GPUs with 80GB of memory that are pairwise NVLinked. We use Jax 0.4.4 with cuda 11.0, cudnn 8.2, and Pytorch 1.8.1 for data-loading. For inference benchmarks (model speed), unless otherwise stated, our models are 6 layers with a FFN size of 512 with 4 attention heads. For model quality benchmarks, refer to the Appendix for a detailed configuration of the model sizes we use. 

\subsection{Model speed}
\label{performance-evaluation}

\begin{figure*}[t]
    \centering
    \includegraphics[width=\textwidth]{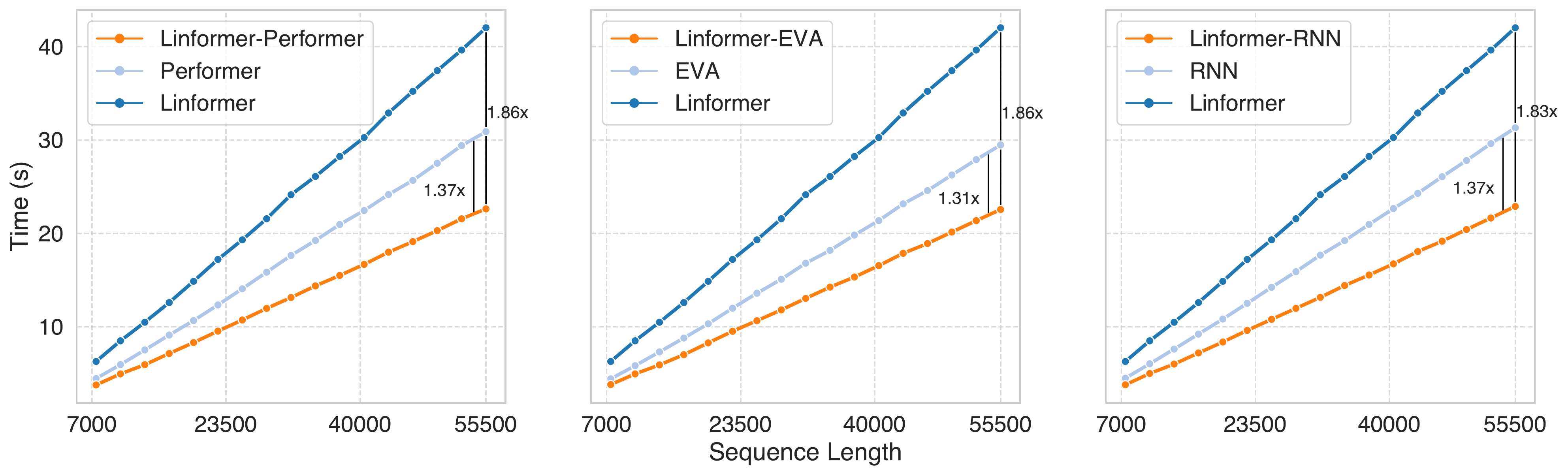}
    \caption{Comparing inference times of all variants as sequence lengths increase. All the models have the same parameter count and were run for 100 iterations of inference. Our method (\name{}) are in green, low-rank is in blue \& kernel is in orange. Lower is better.}
    \label{fig:three-cmp}
\end{figure*}

\begin{table*}[t]
    \centering
    \scalebox{0.81} {
    \begin{tabular}{c|l|ccccccccc|c}
        \toprule
        &Model & COLA & SST-2 & MRPC & STS-B & QQP & \multicolumn{2}{c}{MNLI} & QNLI & RTE & Average  \\
        && (m) & (a) & (f1/a) & (p/s) & (f1/a) & \multicolumn{2}{c}{(a)} & (a)  & (a) \\
        \midrule
         &Full Attention & 48.9 & 90.7 & 89.1/84.8 & 85.5/85.6 & 86.4/90.0 & 81.3 & 81.8 & 88.2 & 65.7 & 81.5\\ 
         \hline
         \midrule 
         \multirow{4}{*}{\rotatebox[origin=c]{90}{\textbf{X-formers}}}&Linformer & 29.2 & 90.0 & 86.3/79.1 & 82.4/82.5 & 84.1/88.1 & 76.9 & 76.4 & 83.9 & 59.9 & 76.6\\
         &Performer & 39.0 & 90.3 & 85.1/76.9 & 81.2/81.1 & 84.7/88.6 & 78 & 77.5 & 84.5 & 62.0 & \textbf{77.4} \\ 
         &RNN & 39.9 & 89.2 & 84.3/76.4 & 81.3/81.4 & 84.4/88.3 & 77.4 & 76.6 & 83.4 & 60.6 & 76.9 \\ 
         &EVA & 51.5 & 91.1 & 82.9/73.8 & 72.8/72.9 & 83.9/88.0 & 75.4 & 76.1 & 80.2 & 59.2 & 75.7 \\ 
         \hline
         \midrule
         \multirow{3}{*}[0.4ex]{\rotatebox[origin=c]{90}{\textbf{\name{}}}}&Lin.-Perf. & 44.1 & 89.9 & 85.9/78.6 & 80.9/81.0 & 84.1/88.1 & 75.6 & 76.2 & 82.4 & 61.3 & \underline{77.3}\\ 
         &Lin.-RNN & 45.3 & 91.0 & 82.3/72.7 & 78.9/79.0 & 84.3/88.3 & 77.4 & 77.1 & 83.4 & 60.6 & 76.7 \\ 
         &Lin.-EVA & 38.7 & 89.1 & 83.2/75.4 & 79.2/79.2 & 83.7/87.9 & 75.7 & 75.6 & 82.2 & 64.6 & 76.2\\
         \bottomrule
    \end{tabular}}
\vspace{2pt}
\caption{A comparison of the GLUE scores in between the full-attention, low-rank (linformer), kernel methods and \name{}. MNLI has two tasks: matched ($6^\textrm{th}$ column) and mismatched ($7^\textrm{th}$ column).}
\label{table:glue-cmp}
\end{table*}

\textbf{Impact of Increasing Sequence Length.} We set $d_m (=2600) > d_k (=1500) > d_h (=325)$, with $H$ (number of heads) to 8 following claim \ref{claim:one}. We vary $N$ from 7.05k to 55.5k in increments of 3k. These are realistic scenarios of parameter configurations. For example, GPT-4 has a sequence length of 32k \citep{gpt4}, and Claude has a sequence length of 100k \citep{claude}. Figure \ref{fig:three-cmp} shows the result. As the sequence length increases, the speedups of \name{} over low-rank and kernel methods increase. This culminates in the linformer-performer variant experiencing a 1.86x and 1.37x, the linformer-RNN variant experiencing a 1.83x and 1.37x, and the linformer-EVA variant experiencing a 1.86x and 1.31x speedup over each of its respective constituent low-rank and kernel method.

\textbf{Speedups over full-attention and flash-attention.} The speedups attained over low-rank and kernel methods are non-trivial and \textit{compound} against full-attention. Figure \ref{fig:speed-ablations} (top) and \ref{fig:speed-ablations} (bottom) illustrate this compounding effect with \name{} variants experiencing speedups of up to 25.1x over full-attention. Moreover, \name{} experience speedups of up to 23x over SOTA flash-attention \cite{flash-attention}. See Appendix C for our evaluation against flash-attention.

\textbf{Model speed discussion.} Intuitively, these speedups occur because \name{} inherit the performance strengths of both underlying kernel and low-rank methods. Speedups against kernel methods occur because the hidden dimension is large enough such that the following operations: (1) linear transformations and (2) $KV$ products contribute significantly to run-time. \name{}, in applying low-rank approximation prior to these operations cut down the number of keys and values and accordingly massively reduce the cost of this product. 

Speedups against low-rank methods occur because the $d_k$ \& $N$ are large enough such that the following operations: (1) softmax on $A_i$ and (2) $QK^T$ \& $A_iV$ products contribute significantly to runtime. \name{}, in cheaply approximating the softmax as well as re-ordering the matrix multiplications offset some of this cost. 

\subsection{Model quality}
\label{accuracy-evaluation}

We train our models across a variety of tasks: language modeling (Wikitext-103), language understanding (GLUE), long sequence modeling (LRA), machine translation (English to German and English to French), and image classification (ImageNet). See Appendix for a detailed setup of the hyperparameters we use.

\textbf{Masked Language Modeling.} We pre-train with the masked language modeling objective on wikitext-103 for 120k steps. Our results are in table \ref{table:collective-tasks}. The linformer-EVA variant outperforms linformer by 5.5\% and is within 13\% of EVA, while the linformer-RNN variant outperforms linformer by 5\% and is within 0.8\% of RNN. The linformer-performer variant is within 2\% of linformer \& 9\% of performer's perplexity scores. 

\textbf{Language Understanding.} We fine-tune on GLUE after pre-training on wikitext-103. Our results are in table \ref{table:collective-tasks}. The linformer-performer variant outperforms linformer by 0.9\% and is within 0.1\% of performer. The linformer-RNN variant outperforms linformer by 0.2\% and is within 0.2\% of RNN. The linformer-EVA variant surpasses EVA by 0.6\% and is within 0.5\% of linformer.

\textbf{Long Sequence Modeling.} We train on the LRA benchmark following the setup in \cite{skyformer}. Our results are in table \ref{table:lra-cmp}. The linformer-performer variant is within 2.9\% of linformer and 7.9\% of performer. The linformer-RNN variant outperforms linformer by 0.05\% and is within 3.9\% of RNN. The linformer-EVA variant outperforms linformer by 2.7\% and is within 4.6\% of EVA.   

\begin{table*}[]
    \centering
    \scalebox{0.85} {
    \begin{tabular}{c|l|ccccc|c}
        \toprule
        &Model & ListOps & Text & Retrieval & Image & Path & Average  \\
        \midrule
         &Full Attention & 37.95 & 60.18 & 80.45 & 37.39 & 70.19 & 57.23\\ 
         \hline
         \midrule 
         \multirow{4}{*}{\rotatebox[origin=c]{90}{\textbf{X-formers}}}&Linformer & 37.85 & 55.80 & 78.61 & 37.72 & 64.75 & 54.95\\
         &Performer & 38.51 & 59.62 & 80.77 & 38.14 & 71.24 & \underline{57.66}\\ 
         &RNN & 37.20 & 64.98 & 76.92 & 38.19 & 68.58 & 57.17\\ 
         &EVA & 38.61 & 65.16 & 81.00 & 41.60 & 69.27 & \textbf{59.13} \\ 
         \hline
         \midrule
         \multirow{3}{*}[0.4ex]{\rotatebox[origin=c]{90}{\textbf{\name{}}}} & Lin.-Perf. & 37.55 & 57.61 & 65.82 & 39.30 & 66.68 & 53.40 \\ 
         &Lin.-RNN & 36.84 & 58.46 & 75.72 & 37.60 & 66.27 & 54.98 \\ 
         &Lin.-EVA & 37.30 & 60.15 & 74.35 & 42.29 & 68.48 & 56.51\\
         \bottomrule
    \end{tabular}}
\vspace{3pt}
\caption{A comparison of the LRA scores in between the full-attention, low-rank (linformer), kernel methods and \name{}.}
\label{table:lra-cmp}
\end{table*}

\textbf{Auto-regressive Machine Translation.} We train on the english to german and english to french machine translation tasks for 128k steps using the t5x\footnote{https://github.com/google-research/t5x} library. Our results are in table \ref{table:collective-tasks}. We observe that the linformer-EVA variant outperforms linformer, performer, RNN \& EVA by at least 1\%. The linformer-performer variant outperforms performer by 8.5\% and is within 0.3\% of linformer's BLUE. The linformer-RNN variant outperforms RNN by 0.9\% and is within 6\% of linformer's BLUE. 

\textbf{Image Classification.} We train a T2T-ViT \cite{t2tvit} on imagenet for 350 epochs. Our results are in table \ref{table:collective-tasks}. We observe that the linformer-EVA variant outperforms linformer, performer, RNN \& EVA by at least 0.4\% across both top-1 and top-5 accuracy. The linformer-performer variant outperforms performer by 0.01\% (top-1) and 0.009\% (top-5) and is within 3\% (top-1) and 2\% (top-5) of linformer. The linformer-RNN variant is within 0.02\% (top-1) and 0.1\% (top-5) of RNN as well as within 3.8\% (top-1) and 1\% (top-5) of linformer.  

\textbf{Model quality discussion.} Low-rank and kernel methods exhibit high model quality across different tasks with \name{} unifying their strengths. Our results indicate that low-rank methods are good at image classification and machine-translation, while kernel methods are good at language understanding and masked language modeling. Nevertheless, across all 6 workloads (GLUE, wikitext-103, LRA, $\texttt{En} \rightarrow \texttt{De}$, $\texttt{En} \rightarrow \texttt{Fr}$ and imagenet) and 3 \name{} variants resulting in 18 data-points, 12 rank between their underlying low-rank and kernel approximations, 3 rank better than both, while 3 rank worse than both. Moreover, most of the 12 data points that rank between the quality of underlying low-rank and kernel approximations closely approach the better of the two. 

\begin{table*}[t]
\centering
\scalebox{0.85} {
\begin{tabular}{c|l|c|c|c|c|c}
  \toprule
  &  & \multicolumn{2}{c|}{NMT} & \multicolumn{2}{c|}{ImageNet} & Wikitext-103 \\\cline{3-7}
  &Model & $\texttt{En} \rightarrow \texttt{De}$ & $\texttt{En} \rightarrow \texttt{Fr}$ & Top-1 & Top-5 & Perplexity\\
  \midrule
  &Full Attention & 24.65 & 32.15 & 71.662 & 90.982 & 5.533 \\
  \hline
  \midrule
  \multirow{4}{*}{\rotatebox[origin=c]{90}{\textbf{X-formers}}}&Linformer & \underline{24.37} & 31.82 & 70.272 & 89.802 & 7.696\\
  &Performer & 22.38 & 29.99 & 67.52 & 88.144 & \underline{7.206} \\
  &RNN & 22.52 & 29.36 & 67.43 & 87.98 & 7.228 \\
  &EVA & 24.16 & \underline{32.12} & \underline{70.322} & \underline{89.994} & \textbf{6.432} \\
  \hline
  \midrule
  \multirow{3}{*}[0.4ex]{\rotatebox[origin=c]{90}{\textbf{\name{}}}}&Linformer-Performer & 24.30 & 31.21 & 67.592 & 88.23 & 7.867 \\
  &Linformer-RNN & 22.72 & 29.73 & 67.416 & 88.106 & 7.293 \\
  &Linformer-EVA & \textbf{24.61} & \textbf{32.37} & \textbf{70.662} & \textbf{90.366} & 7.271 \\
  \bottomrule
\end{tabular}}
\vspace{3pt}
\caption{A comparison between all the models on 3 tasks: machine translation, image classification and masked language modeling. $\texttt{En} \rightarrow \texttt{De}$ and $\texttt{En} \rightarrow \texttt{Fr}$ are the English to German and English to French translation tasks respectively.}
\label{table:collective-tasks}
\end{table*}

Despite theorem \ref{accuracy-claim} indicating that \name{}'s approximation error is the sum of underlying low-rank and kernel approximations, empirically \name{}'s model quality approaches the better of its constituent models. In practice, the gap in theorem \ref{accuracy-claim} is mitigated by up-training \cite{sparse-upcycling, gqa}, where an $\alpha$ fraction of the training steps trains either a low-rank or kernelized base model, thereafter training a \name{} variant for the remaining $1-\alpha$ fraction of steps whose parameters are initialized with the base model. Our ablations in section \ref{sub-section:ablations} indicate that up-training averages the approximation error of underlying low-rank and kernel approximations. Moreover, tuning $\alpha$ results in \name{} variants whose quality approaches the better of the two. 

\subsection{Ablations}
\label{sub-section:ablations}

We conduct two sets of ablations studies investigating the impact of hyperparameters on (1) model speed (see section \ref{subsubsection-run-time-perf}), and (2) model quality (see section \ref{subsubsection-accuracy}). Our model speed ablations ascertain the impact of the downsampling factor (a hyperparameter introduced by low-rank approximation) and hidden dimension on step-time. Our model quality ablations ascertain the impact of up-training.

\subsubsection{Model speed}
\label{subsubsection-run-time-perf}

\begin{figure}
    \centering
    \includegraphics[width=0.96\textwidth]{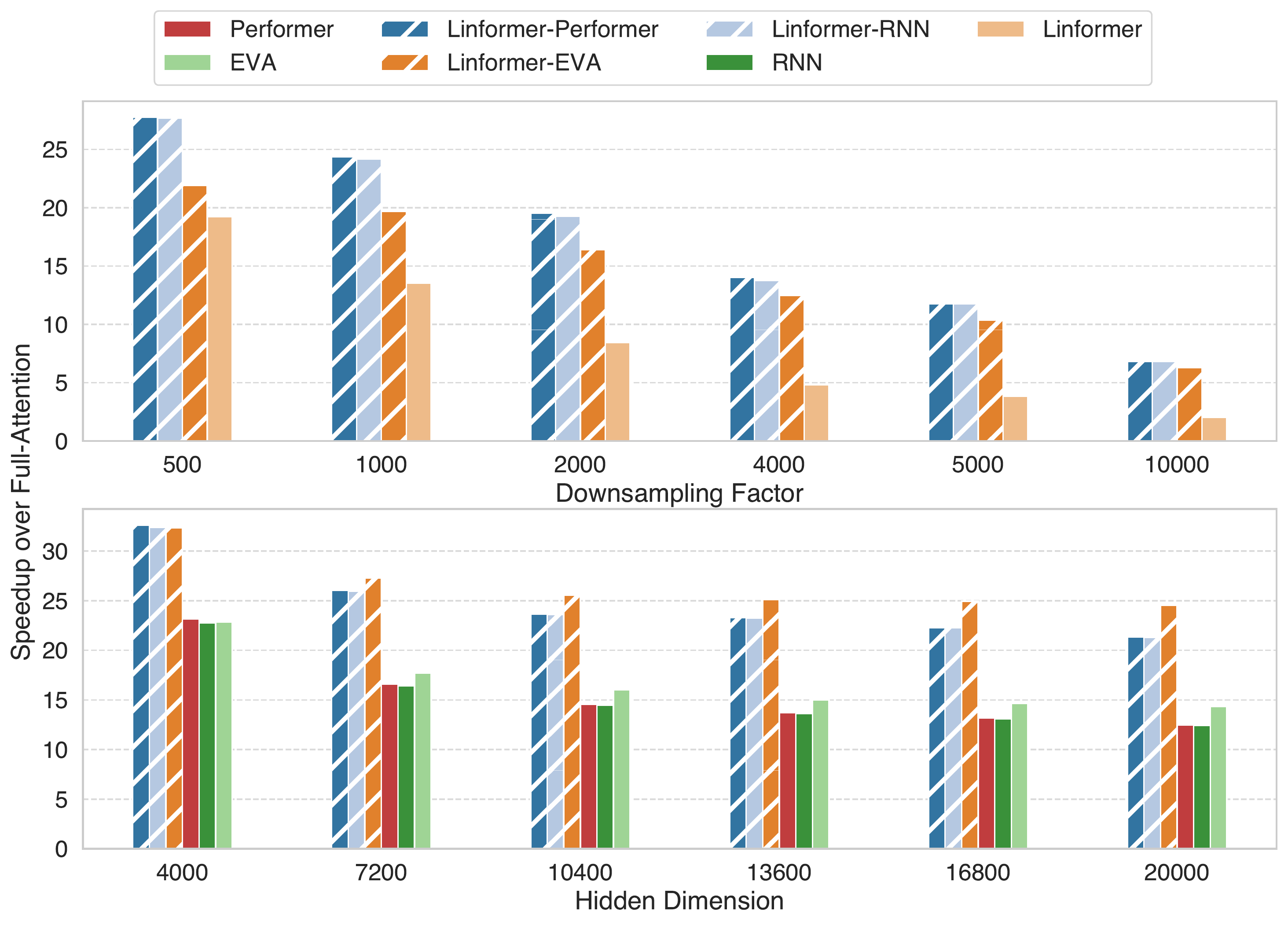}
    \caption{The impact of the downsampling factor and hidden dimension on runtime performance (model speeds) normalized to speedups over full-attention. The top figure compares \name{} to low-rank methods and the bottom figure compares \name{} to kernel methods. Our methods (\name{}) are highlighted in hashed bars, while low-rank and kernel methods are highlighted in solid bars.}
    \label{fig:speed-ablations}
\end{figure}

\textbf{Impact of Increasing Downsampling Factor.} We set $N (=20k) > d_k > d_h (=128)$. We vary $d_k$ from 8k to 20k in increments of 3.2k. The sequence lengths and head hidden dimension are similar to other SOTA models. Figure \ref{fig:speed-ablations} (top) shows the result normalized to speedups over full-attention. As the downsampling factor increases, the speedup of all the \name{} over low-rank magnifies. This culminates in the linformer-performer variant incurring a 3.39x, the linformer-RNN variant incurring a 3.38x, and the linformer-EVA variant incurring a 3.12x speedup over linformer (low-rank). 

\textbf{Impact of Increasing Hidden Dimension.} We set $N (=20k) > d_k(H+2) (=6k), d_m > d_k (=1k)$. We vary $d_m$ from 800 to 20k in increments of 3.2k. Figure \ref{fig:speed-ablations} (bottom) shows the result normalized to speedups over full-attention. As the hidden dimension increases, the speedup of all the \name{} over their kernel methods intensifies. This culminates in the linformer-performer variant enjoying a 1.72x, the linformer-RNN variant enjoying a 1.72x and the linformer-EVA variant enjoying a 1.71x speedup over each of its kernel methods respectively. 

\subsubsection{Model quality}
\label{subsubsection-accuracy}
\begin{table}
\centering
\scalebox{0.85} {
\begin{tabular}{cc|c|c|c|c|c|c|c|c}
  \toprule
  \multicolumn{2}{c|}{Kernel} & Low-rank & \multicolumn{7}{c}{\name{}} \\\cline{4-10} \noalign{\vspace{0.25ex}}
  & & & 12.5\% & 25\% & 37.5\% & 50\% & 62.5\% & 75\% & 87.5\%  \\ 
  \midrule
   \multicolumn{1}{c|}{Performer} & 27.97 & \multirow{3}{*}{28} & 27.98 & 27.68 & \textbf{28.00} & 27.86 & 27.86 & 27.56 & 27.66\\
   \multicolumn{1}{c|}{RNN} & 23.41 & & 23.94 & \textbf{24.57} & 24.23 & 24.42 & 24.34 & 24.25 & 23.45 \\
   \multicolumn{1}{c|}{EVA} & 28.54 & & 28.70 & 28.72 & \textbf{28.87} & 28.65 & 28.73 & 28.66 & 28.64  \\
  \bottomrule
\end{tabular}}
\vspace{3pt}
 \caption{We compare \name{} variants up-trained on different $\alpha$ from 12.5\% to 87.5\% to constituent low-rank and kernel methods on the english to german machine-translation task. The first, second, and third rows correspond to the linformer-performer, linformer-RNN, and linformer-EVA variants, respectively. We report the BLEU scores of each run.}
 \label{table:up-train}
\end{table}

\textbf{Impact of up-training on model quality.} We up-train \name{} with different ratios of $\alpha$ (percent of the training steps spent on the base model) and compare their quality to constituent kernel and low-rank methods. We use the english to german machine-translation task. Our results are in table \ref{table:up-train}. We note 3 observations. (1) Majority of the up-trained \name{} variants, 16 out of 21, rank between constituent low-rank and kernel methods. (2) The highest quality linformer-performer and linformer-EVA variants outperform all the kernel and low-rank methods. (3) The highest quality up-trained \name{} variants have ratios of $\alpha$ within the [25\%, 37.5\%] range. These observations imply that up-training averages the approximation error of constituent low-rank and kernel methods, producing \name{} variants whose quality approaches the higher-quality constituent model. Moreover, the optimal $\alpha$ ratios are low, resulting in a small fraction of time spent training the base-model.

\section{Conclusion}
We propose a new technique to unify low-rank and kernel methods, producing a family of transformers, \name{}. \name{} are fast, incurring end-to-end speedups of up to 1.7x and 3.3x over kernel and low-rank methods respectively. Moreover, they are of high quality across diverse tasks. Across 6 benchmarks ranging text and image modalities \name{} are competitive with, and occasionally surpass, the underlying low-rank and kernel methods that compose them. As a result of their training-efficient nature, \name{} require fewer FLOPs to attain similar loss levels compared to full-attention, reducing the computational costs of training high-quality models for a diverse range of tasks.

\section*{Acknowledgements}
We would like to thank Hao Peng for introducing us to the EVA model, Krzysztof Choromanski for his help in how to train transformers to achieve SOTA results, Tianle Cai for his insight on how to produce a general theoretical claim, as well as Minjia Zhang and Wanyu Zhao for their feedback on the draft of the paper. This work is in part supported by ACE, one of the seven centers in JUMP 2.0, a Semiconductor Research Corporation (SRC) program sponsored by DARPA and generous support from Google Cloud Credits. 

\nocite{*}
\bibliography{bibliography}

\newpage


\appendix
\section*{Appendix}
\newtheoremstyle{named}{}{}{\itshape}{}{\bfseries}{.}{.5em}{\thmnote{#3's }#1}
\theoremstyle{named}
\newtheorem*{namedtheorem}{Theorem}

\graphicspath{ {./figures/} }

\newtheorem{definition}{Definition}

\section{Training Setup}
\subsection{Metrics}
We measure the following for all our models:
\begin{itemize}
    \item Perplexity: we measure the perplexity across pre-training on Wikitext-103 and BookCorpus.
    \item Accuracy: for certain tasks in GLUE.
    \item F1: for the QQP and MRPC tasks in GLUE.
    \item Matthews Correlation: for the COLA task in GLUE.
    \item Pearson \& Spearman correlation: for the STS-B task in GLUE.
    \item BLEU: for the neural machine translation tasks in WMT'14.
\end{itemize}

\subsection{Datasets}

\textbf{Wikitext-103} Wikitext-103 is a collection of good and verified articles from Wikipedia. It consists of over 100 million tokens and is a popular dataset to pre-train self-attention models on.

\textbf{GLUE} The General Language Understanding Evaluation (GLUE) is a collection of 9 tasks that evaluate natural language understanding systems. It is a popular benchmark to fine-tune self-attention models on.  

\textbf{LRA} Long Range Arena (LRA) is a task suite that evaluates transformer model qualities under long-context scenarios. It consists of 6 tasks that cover multiple modalities including text, and image.

\textbf{ImageNet} ImageNet is a collection of over 14 million annotated images according to the WordNet hierarchy. It is a popular dataset for image and object classification.

\subsection{Model Configurations}
In MLM, NLU, LRA, and image classification, we use encoder-only architectures. In NMT and model quality ablations, we use encoder-decoder architectures with our techniques applied to the encoder only. A detailed setup of each model's configuration is shown in table \ref{table:settings}.

\subsection{Up-training Ratios}
For NMT, the linformer-performer and linformer-EVA variants are up-trained with a ratio of $\alpha=80\%$ with a low-rank base model. Linformer-RNN is up-trained with a ratio of $\alpha=64\%$ with a kernel base model. For image classification, linfromer-performer and linformer-EVA are up-trained with a ratio of $\alpha=16.7\%$ with a low-rank base model. Linformer-RNN is up-trained with a ratio of $\alpha=33.3\%$ with a kernel base model. No up-training is applied to any other experiment.

\begin{table*}[h]
    \centering
    \scalebox{0.78} {
\begin{tabular}{ll|ccc|cccc|c}
\toprule
\multicolumn{2}{c|}{Task} & Batch Size & LR & LR Schedule  & \#Layers & \#Heads & Head Dim & FFN Dim & \#Params \\
\hline
\midrule
\multicolumn{2}{l|}{MLM} & \multirow{2}{*}{1024} &  1e-4,1.2e-4 & \multirow{2}{*}{linear decay} &  \multirow{2}{*}{12} &  \multirow{2}{*}{12} & \multirow{2}{*}{64} & \multirow{2}{*}{3072} & \multirow{2}{*}{108M}   \\
\multicolumn{2}{l|}{NLU} & & [1e-4,5e-4] &  &  &   &  &  &  \\
\midrule
\multicolumn{1}{l|}{\multirow{5}{*}{LRA}} & ListOps & 32 & 1e-4 & \multirow{5}{*}{linear decay}  & \multirow{5}{*}{2} & \multirow{5}{*}{2} & \multirow{5}{*}{64} & \multirow{5}{*}{128} & \multirow{5}{*}{0.4M}  \\
\multicolumn{1}{l|}{} & Text & 32 & 1e-4 &  &   &  &  &  &  \\
\multicolumn{1}{l|}{} & Retrieval & 16 & 2e-4 &   &  &  &  &  &  \\
\multicolumn{1}{l|}{} & Image & 256 & 1e-4 &  &  &   &  &  &  \\
\multicolumn{1}{l|}{} & Pathfinder & 128 & 2e-4 &  &   &  &  &  &  \\
\midrule
\multicolumn{2}{l|}{NMT} & 128 & 1 & inverse square root & 12 & 12 & 64 & 3072 &  220 M \\ 

\multicolumn{2}{l|}{Image Classification} & 128 & 1e-3 & cosine anneal  & 7 & 4 & 64 & 512 & 4.3M \\

\multicolumn{2}{l|}{Quality Ablations} & 128 & 1 & inverse square root &  6 & 8 & 64 & 2048 & 60.5 M\\
\bottomrule

\end{tabular}
    
    }
\caption{Training setting and model size used in evaluation section.}
\label{table:settings}
\end{table*}

\section{Efficiency Analysis}

We analyze where \name{} incurs fewer FLOPs compared to low-rank methods, kernel methods as well as both \textit{at the same time}. 
\begin{claim}
    For sequence length: $N$, downsampling factor: $d_k$, head hidden dimension: $d_h$ when: $$N-1 > d_k> d_h $$ \name{} incur lower FLOPs against low-rank methods. 
    \label{claim:two}
\end{claim}
\begin{claim}
    For sequence length: $N$, hidden dimension: $d_m$, downsampling factor: $d_k$, number of heads: $H$, when: $$N > d_k(H+2), d_m > d_k$$ \name{} incur lower FLOPs against kernel methods.
    \label{claim:three}
\end{claim}
Each of the above claims delineates a \textit{regime} where \name{} exhibit competitive performance over low-rank and kernel techniques. Moreover, each bound is tight, requiring sequence lengths larger than 1k to imply speedups. 

Our argument for all the claims is similar. We first compute the runtime of \name{}, kernel and low-rank methods by breaking up the steps of MHSA into the following:
\begin{itemize}
    \item The low-Rank downsampling products: $E_1K$ \& $E_2V$ (low-rank \& \name{} only) 
    \item  Linear Transformations: $Q'=QW^Q$, $K'=KW^K$ \& $V'=VW^V$.
    \item The product: $Q'K'^T$ (low-rank only)
    \item The application of the kernel: $\phi(Q')$ \& $\phi(K')$ (\name{} \& kernel only)
    \item The softmax (low-rank only)
    \item The product $AV'$ (low-rank only). Or the product $Q'(K'^TV')$ (\name{} \& kernel only).
\end{itemize}

We can sum the time across all the steps of MHSA for each method and compare when one method's expression is greater than the others'. Throughout our arguments, we assume that the kernel $\phi$ that parameterises a kernel method produces vectors whose dimensionality is on the same order of magnitude as the hidden dimension. We use the following notation for convenience: $H$ - number of heads, $d_k$ - downsampling factor, $d_m$ - hidden dimension, $N$ - sequence length, and $d_h$ - head hidden dimension. 

\textbf{Cumulative runtime of low-rank MHSA}. The time taken for the downsampling step: $$2Nd_kd_m$$ The time taken for the Linear transformation step: $$Nd_m^2 +2d_kd_m^2$$ The time taken for the $QK^T$ product: $$Nd_md_k$$ The time taken for the softmax: $$NHd_k$$ The time taken for the $AV$ product: $$Nd_kd_m$$ With the total time taken as: 
\begin{equation}
    2Nd_kd_m + Nd_m^2 +2d_kd_m^2 + Nd_md_k + NHd_k + Nd_kd_m
    \label{low-rank-cum}
\end{equation}

\textbf{Cumulative runtime of kernel MHSA}. The time taken for the Linear Transformation step: $$3Nd_m^2$$ The time taken for the application of the kernel: $$2Nd_m$$ The time taken for the application of the $Q(K^TV)$ product: $$2Nd_md_h$$ With the total time taken as: 
\begin{equation}
    3Nd_m^2 + 2Nd_m + 2Nd_md_h
    \label{kernel-cum}
\end{equation}

\textbf{Cumulative runtime of \name{}' MHSA}. The time taken for the downsampling step: $$2Nd_kd_m$$ The time taken for the linear transformation step: $$Nd_m^2 + 2d_kd_m^2$$ The time taken for the application of the kernel is: $$Nd_m + d_md_k$$ The time taken for the $QK^TV$ product is: $$d_kd_md_h + Nd_hd_m$$ With the total time taken as: 
\begin{equation}
    2Nd_kd_m + Nd_m^2 + 2d_kd_m^2 + Nd_m + d_md_k + d_kd_md_h + Nd_hd_m
    \label{swift-cum}
\end{equation}

\subsection{Argument for Claim 1}

To show claim 1 - when \name{} incur lower FLOPs over both low-rank and kernel methods, we have to that $N>d_k(H+2)>d_m>d_k>d_h \rightarrow$ \ref{low-rank-cum} - \ref{swift-cum} $>$ 0 $\wedge$ \ref{kernel-cum} $-$ \ref{swift-cum} $>$ 0. 

We first subtract the runtime of \name{} from the runtime of low-rank methods to produce the following expression: 
\begin{equation}
    \underbrace{Nd_md_k}_{a} + \underbrace{NHd_k}_{b} + \underbrace{Nd_kd_m}_{c} - \underbrace{Nd_m}_{d} - \underbrace{d_md_k}_{e} - \underbrace{d_kd_md_k}_{f} - \underbrace{Nd_hd_m}_{g}
    \label{swift-lowrank} 
\end{equation}
We see that: $d_h < d_k\rightarrow b > d \wedge c > g$ and $N>d_m>d_k \rightarrow N > 1+d_k \rightarrow a > e + f$. We thus have that: when $N>1+d_k,d_k>d_h$ that $a+b+c > d+e+f+g$. 

Next, we subtract the runtime of \name{} from the runtime of kernel methods to get: 
\begin{equation}
    \underbrace{2Nd_m^2}_{i} + \underbrace{Nd_m}_{j} + \underbrace{2Nd_md_h}_{k} - \underbrace{2Nd_kd_m}_{l} - \underbrace{2d_kd_m^2}_{m} - \underbrace{d_md_k}_{n} - \underbrace{d_kd_md_h}_{o}
    \label{swift-kernel}
\end{equation}
We see that $d_m > d_k \rightarrow i > l$. We next collectively consider when $j+k>m+n+o$. This occurs when:
\begin{align*}
    Nd_m + 2Nd_md_h - 2d_kd_m^2 - d_md_k - d_kd_md_h > 0 \\
    \rightarrow d_m(N + 2Nd_h - 2d_kd_m - d_k - d_kd_h) > 0 \\
    \rightarrow N + 2Nd_h - 2d_kd_m - d_k - d_kd_h > 0 \\
    \rightarrow N(1+2d_h) - d_k(2d_m + 1 + d_h) > 0 \\
    \rightarrow N > \underbrace{\frac{d_k(2d_m+1+d_h)}{1+2d_h}}_{z}
\end{align*}
Now, we can simplify $z$ as follows:
\begin{align*}
    &\text{    } \frac{d_k(2d_m+1+d_h)}{1+2d_h} \\
    &= \frac{d_k(2Hd_h+1+d_h)}{1+2d_h} \\
    &= \frac{d_k(2(H+1)d_h+1)}{1+2d_h} \\
    &< \frac{d_k(2(H+1)d_h+1)}{2d_h} \\
    &<  \frac{d_k(2(H+1)d_h+d_h)}{2d_h} \\
    &= \frac{d_k(2(H+2))}{2} = d_k(H+2)
\end{align*}
To finally attain: $N > d_k(H+2) > z \rightarrow j+k>m+n+o$. We thus have when $N>d_k(H+2), d_m>d_k$ that $i+j+k>l+m+n+o$.

It follows that when $N>d_k(H+2)>d_m>d_k>d_h$ \name{} incur lower FLOPs over kernel and low-rank methods. 

\subsection{Argument for Claim 2}

Claim 2 - when \name{} incur lower FLOPs over low-rank - naturally follows from \ref{swift-lowrank} since $N-1 > d_k> d_h \rightarrow$ \ref{low-rank-cum} - \ref{swift-cum} $>$ 0. 

\subsection{Argument for Claim 3}

Claim 3 - when \name{} incur lower FLOPs over kernel methods - naturally follows from \ref{swift-kernel} since $N>d_k(H+2), d_m>d_k \rightarrow$ \ref{kernel-cum} - \ref{swift-cum} $>$ 0.

\section{Comparison against flash-attention}

\begin{figure}[h]
    \centering
    \includegraphics[width=\textwidth]{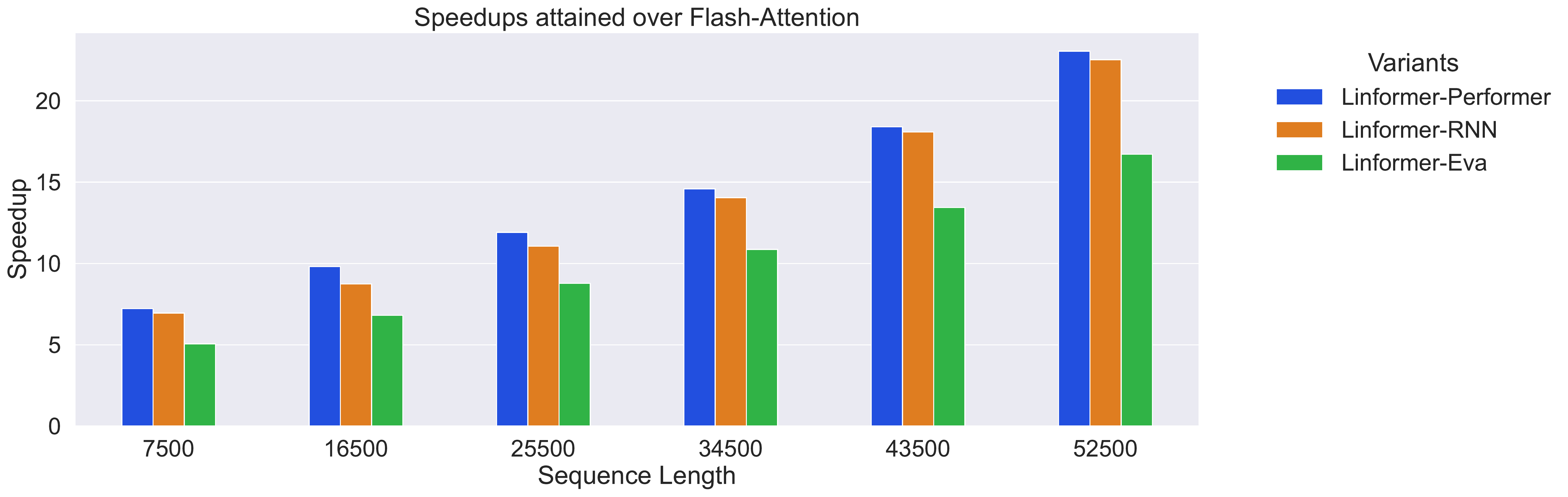}
    \caption{Speedups \name{} attain over Flash-Attention.}
    \label{fig:fmha_speedups}
\end{figure}

We compare the step-times of \name{} variants against SOTA flash-attention's implementation. We use a 6 layer transformer, with a FFN hidden dimension of 512 and 4 attention heads. We vary the sequence length from 7500 to 52500 in increments of 9000. Our results are in figure \ref{fig:fmha_speedups}. Across all sequence lengths, \name{} yield considerable speedups over flash-attention, culminating in speedups of up to 23x. 

\section{Accuracy Analysis}

\subsection{Proof of Theorem 1}

We first define important structures to be used throughout this proof.

\begin{definition}
In computing canonical multi-head self-attention, suppose we have the queries, keys and values Q,K,V $\in \mathbb{R}^{n\times d_m}$ respectively. As well as the learnt weights: $W_i^Q$, $W_i^K$ and $W_i^V \in \mathbb{R}^{n \times d_h}$. We have that: $$A_i = softmax(\frac{QW^Q_i(KW_i^K)^T}{\sqrt{d_k}})VW_i^V$$ Furthermore, we denote: $$A_{softin}=\frac{QW^Q_i(KW_i^K)^T}{\sqrt{d_k}}$$
\end{definition}

We also cite the following theorem from Linformer, which will form the basis for our proof:
\begin{namedtheorem}[Linformer]
For any $Q_i$, $K_i$, $V_i$ $\in \mathbb{R}^{n\times d_m}$ and $W_i^Q$, $W_i^K$, $W_i^V$ $\in \mathbb{R}^{d_m \times d_h}$.  If $k = 5\log(d)/(\epsilon_2^2 - \epsilon_3^2)$ where $\epsilon_2$ and $\epsilon_3$ are defined as per Linformer Theorem 2, and we further define the matrices $E_i = \delta R, F_i = e^{-\delta}R$ where $R \in \mathbb{R}^{n \times k}$ whose entries are iid sampled from $N(0, 1/k)$. We have that, for any row-vector $w \in A_{softin}$: $$Pr(\lVert softmax(wE_i^T)F_iVW_i^V  - softmax(w)VW_i^V\lVert_{2} < \epsilon_1 \lVert softmax(w)\lVert_{2} \lVert VW_i^V\lVert_{2}) > 1 - o(1) $$
\end{namedtheorem}

We re-write the above into a form that simplifies our proof: 

\begin{equation}
    Pr(\lVert softmax(A_{softin}E_i^T)F_iVW_i^V  - A_iVW_i^V\lVert_{\infty} < \epsilon_1 \lVert A_i\lVert_{2} \lVert VW_i^V\lVert_{2}) > 1 - o(1)
    \label{linformer}
\end{equation}

Now, let us assume we have a random-feature based kernel method parameterised by $\phi(x) = \frac{1}{\sqrt{m}}[\psi_1(x), ... \psi_m(x)]$ such that: $$\mathbb{E}[\psi_i(x)^T\cdot \psi_i(y)] = \exp(x^T \cdot y)$$ 

We have the following: $$\phi(x)^T \cdot \phi(y) = \frac{1}{m}\sum_{i=1}^{m}\psi_i(x)^T \cdot 
\psi_i(y)$$ 

We then have, by the law of large numbers: 
\begin{align*}
&\lim_{m\rightarrow \infty} \phi(x)^T\cdot \phi(y)\\
&= \lim_{m \rightarrow \infty} \frac{1}{m}\sum_{i=1}^{m}\psi_i(x)^T \cdot \psi_i(y)\\ 
&= \mathbb{E}[\psi_i(x)^T \cdot \psi_i(y)] \\
&= \exp(x^T \cdot y)\\
\end{align*}

Establishing that $\lim\limits_{m \rightarrow \infty}\phi(x)^T \cdot \phi(y) = \exp(x^T \cdot y)$. Therefore, by the definition of a limit, for $\epsilon_4 >0$, $\epsilon_5 = (\epsilon_4 F)/2N^2$, $\displaystyle F = \min_{1\leq i \leq N} \abs{\sum_{k=1}^{N}\phi(x_i)^T \cdot \phi(y_k)} $, $\exists m \in \mathbb{Z}^{+}$ such that: \begin{equation}
\abs{\phi(x)^T \cdot \phi(y) - \exp(x^T \cdot y)} < \epsilon_5
\end{equation}
With $x_i \in QW_i^{Q}$ and $y_k \in KW_i^K$. Next, we bound $\lVert \hat{A}_i - A_i \lVert_{\infty}$, where $\hat{A}$ is the attention matrix materialised by the random-feature based kernel method parameterised by $\phi$:
\begin{align*}
    & \lVert \hat{A}_i - A_i \lVert_{\infty} \\
    & = \max_{1 \leq i \leq N} \sum_{j=1}^{N}\abs{a_{ij}} \\
    & = \max_{1 \leq i \leq N} \sum_{j=1}^{N}\abs{\frac{\phi(x_i)^T \cdot \phi(y_j)}{\sum_{k=1}^{N}\phi(x_i)^T \cdot \phi(y_k)} - \frac{\exp(x_i^T \cdot y_j)}{\sum_{k=1}^{N} \exp(x_i^T \cdot y_k)}} \\
    & = \max_{1 \leq i \leq N}\sum_{j=1}^{N} \abs{\frac{\phi(x_i)^T \cdot \phi(y_j)\sum_{k=1}^{N} \exp(x_i^T \cdot y_k)-\exp(x_i^T \cdot y_j)\sum_{k=1}^{N}\phi(x_i)^T \cdot \phi(y_k)}{\sum_{k=1}^{N}\phi(x_i)^T \cdot \phi(y_k)\sum_{k=1}^{N} \exp(x_i^T \cdot y_k)}}  \\
    & \leq \max_{1 \leq i \leq N}\sum_{j=1}^{N} \abs{\frac{[\exp(x_i^T \cdot y_j) + \epsilon_5]\sum_{k=1}^{N} \exp(x_i^T \cdot y_k)-\exp(x_i^T \cdot y_j)\sum_{k=1}^{N}[\exp(x_i^T \cdot y_k) + \epsilon_5]]}{\sum_{k=1}^{N}\phi(x_i)^T \cdot \phi(y_k)\sum_{k=1}^{N} \exp(x_i^T \cdot y_k)}}  \\
    & = \max_{1 \leq i \leq N} \sum_{j=1}^{N} \abs{\frac{\epsilon_5\sum_{k=1}^{N} \exp(x_i^T \cdot y_k)- N\epsilon_5\exp(x_i^T \cdot y_j)}{\sum_{k=1}^{N}\phi(x_i)^T \cdot \phi(y_k)\sum_{k=1}^{N} \exp(x_i^T \cdot y_k)}}\\
    & \leq \max_{1 \leq i \leq N} \sum_{j=1}^{N} \abs{\frac{\epsilon_5\sum_{k=1}^{N} \exp(x_i^T \cdot y_k)+ N\epsilon_5\exp(x_i^T \cdot y_j)}{\sum_{k=1}^{N}\phi(x_i)^T \cdot \phi(y_k)\sum_{k=1}^{N} \exp(x_i^T \cdot y_k)}}\\
    & \leq \max_{1 \leq i \leq N} \sum_{j=1}^{N} \abs{\frac{\epsilon_5\sum_{k=1}^{N} \exp(x_i^T \cdot y_k)+ N\epsilon_5\sum_{k=1}^{N}\exp(x_i^T \cdot y_k)}{\sum_{k=1}^{N}\phi(x_i)^T \cdot \phi(y_k)\sum_{k=1}^{N} \exp(x_i^T \cdot y_k)}}\\
    & = \max_{1 \leq i \leq N} \sum_{j=1}^{N} \abs{\frac{\epsilon_5+ N\epsilon_5}{\sum_{k=1}^{N}\phi(x_i)^T \cdot \phi(y_k)}}\\
    & = \max_{1 \leq i \leq N} \sum_{j=1}^{N} \frac{\epsilon_5+ N\epsilon_5}{\abs{\sum_{k=1}^{N}\phi(x_i)^T \cdot \phi(y_k)}}\\
    & \leq \max_{1 \leq i \leq N} \sum_{j=1}^{N} \frac{\epsilon_5+ N\epsilon_5}{\min_{i \leq I \leq N}\abs{\sum_{k=1}^{N}\phi(x_i)^T \cdot \phi(y_k)}}\\
    & = \max_{1 \leq i \leq N} \sum_{j=1}^{N} \frac{\epsilon_4F/2N^2+ \epsilon_4F/2N}{\min_{i \leq I \leq N}\abs{\sum_{k=1}^{N}\phi(x_i)^T \cdot \phi(y_k)}}\\
    & = \max_{1 \leq i \leq N} \sum_{j=1}^{N} \frac{\epsilon_4}{2N^2}+ \frac{\epsilon_4}{2N}\\
    & \leq \max_{1 \leq i \leq N} \sum_{j=1}^{N} \frac{\epsilon_4}{2N}+ \frac{\epsilon_4}{2N}\\
    & \leq \max_{1 \leq i \leq N} \sum_{j=1}^{N} \frac{\epsilon_4}{N}\\
    & = \epsilon_4
\end{align*}

To get the final bound:
\begin{equation}
    \lVert \hat{A}_i - A_i \lVert_{\infty} < \epsilon_4
\label{one}
\end{equation} 

Substitute $K = E_1K$ in \ref{one} and we get: 

\begin{equation}
 \lVert \phi(QW_i^Q)\phi(E_1KW_i^K)^T - softmax(QW_i^Q(E_1KW_i^K)^T)\lVert_{\infty} < \epsilon_4  
\end{equation}

We can recover the full Linformer style attention computation by right multiplying by the norm $\lVert F_iVW_i^V\lVert_{\infty}$ to obtain:

\begin{equation}
    \lVert \phi(QW_i^Q)\phi(E_1KW_i^K)^T - softmax(QW_i^Q(E_1KW_i^K)^T)\lVert_{\infty}\lVert F_iVW_i^V\lVert_{\infty} < \epsilon_4\lVert F_iVW_i^V\lVert_{\infty}
\label{three}    
\end{equation}

Since $\lVert AB \lVert \leq \lVert A \lVert \lVert B \lVert $, \ref{three} simplifies to: 

\begin{equation}
    \lVert \phi(QW_i^Q)\phi(E_1KW_i^K)^TF_iVW_i^V - softmax(QW_i^Q(E_1KW_i^K)^T) F_iVW_i^V\lVert_{\infty} < \epsilon_4\lVert F_iVW_i^V\lVert_{\infty}
    \label{swift-subtract-linf}
\end{equation}

Now, we add \ref{swift-subtract-linf} to \ref{linformer} to get: 

\begin{multline*}
    \lVert \phi(QW_i^Q)\phi(E_1KW_i^K)^TF_iVW_i^V - softmax(QW_i^Q(E_1KW_i^K)^T) F_iVW_i^V\lVert_{\infty} + \\ \lVert softmax(A_{softin}E_i)F_iVW_i^V  - A_iVW_i^V\lVert_{\infty} < \epsilon_4\lVert F_iVW_i^V\lVert_{\infty} + \epsilon_1 \lVert A_i\lVert_{2} \lVert VW_i^V\lVert_{2}
\end{multline*}

We apply the triangle inequality to the LHS, to simplify the above expression to finally yield:

\begin{equation}
     \lVert \phi(QW_i^Q)\phi(E_1KW_i^K)^TF_iVW_i^V - A_iVW_i^V \lVert_{\infty}  < \epsilon_4\lVert F_iVW_i^V\lVert_{\infty} + \epsilon_1 \lVert A_i\lVert_{2} \lVert VW_i^V\lVert_{2}
\end{equation} With high probability for a carefully chosen $E_i$, $F_i$ and $\phi$.


\end{document}